\documentclass[letterpaper]{bioinfo}
\usepackage[svgnames,dvipsnames]{xcolor}

\copyrightyear{2019} \pubyear{2019}

\newcommand{\deepRAM}{\texttt{deepRAM}}
\newcounter{comments}

\access{Advance Access Publication Date: Day Month Year}
\appnotes{Manuscript Category}
\begin{document}
\firstpage{1}

\subtitle{Subject Section}

\title[deepRAM]{Comprehensive Evaluation of Deep Learning Architectures for Prediction of DNA/RNA Sequence Binding Specificities}
\author[Trabelsi, Chaabane \textit{et~al}.]{Ameni Trabelsi\,$^{ \dagger}$, Mohamed Chaabane\,$^{ \dagger}$ and Asa Ben Hur\,$^{*}$}
\address{Department of Computer Science,
Colorado State University,
Fort Collins, CO 80525, USA}

\corresp{$^\dagger$Denotes equal contribution.\\
$^\ast$To whom correspondence should be addressed.}

\history{Received on XXXXX; revised on XXXXX; accepted on XXXXX}

\editor{Associate Editor: XXXXXXX}

\abstract{\textbf{Motivation:} 
Deep learning architectures have recently demonstrated their power in predicting DNA- and RNA-binding  specificities. 
Existing methods fall into three classes: Some are based on Convolutional Neural Networks (CNNs), others use Recurrent Neural Networks (RNNs), and others rely on hybrid architectures combining CNNs and RNNs.
However, based on existing studies it is still unclear
which deep learning architecture is achieving the best performance. Thus an in-depth analysis and evaluation of the different methods is needed to fully evaluate their relative.\\
\textbf{Results:} In this study, We present a systematic exploration of various deep learning architectures for predicting DNA- and RNA-binding  specificities. For this purpose, we present \deepRAM, an end-to-end deep learning tool that provides an implementation of novel and previously proposed architectures; its fully automatic model selection procedure allows us to perform a fair and unbiased comparison of deep learning architectures.
We find that an architecture that uses k-mer embedding to represent the sequence, a convolutional layer and a recurrent layer, outperforms all other methods in terms of model accuracy. 
Our work provides guidelines that will assist the practitioner in choosing the best architecture for the task at hand, and provides some insights on the differences between the models learned by convolutional and recurrent networks.
In particular, we find that although recurrent networks improve model accuracy, this comes at the expense of a loss in the interpretability of the features learned by the model.\\
\textbf{Availability and implementation:} The source code for \deepRAM{} is available at \href{https://github.com/MedChaabane/deepRAM}{https://github.com/MedChaabane/deepRAM}\\
\textbf{Contact:} \href{asa@cs.colostate.edu }{asa@cs.colostate.edu }\\
\textbf{Supplementary information:} Supplementary data are available at \textit{Bioinformatics}
online.}

\maketitle

\section{Introduction}

DNA- and RNA-binding proteins are involved in many biological processes including transcription, translation, and alternative splicing (\citealt{Ferre,gerstberger}). 
Unfortunately, only some of these binding sites have been identified by biological experiments. Moreover, these experiments are expensive and time-consuming.  
In order to represent binding sites and detect new ones, 
Position Weight Matrices (PWMs) are the most common method to characterize the sequence specificity of a protein thanks to their simplicity and ease of interpretation (\citealt{stormo2000dna}).
However, many studies suggest that sequence specificity can be better captured using more complex models (\citealt{rohs2010origins,kazan2010rnacontext,siggers2013protein}).  

In recent years, deep neural networks have become the technique of choice for challenging tasks in computer vision (\citealt{krizhevsky2012imagenet,lecun2015deep}), speech recognition (\citealt{hinton2012deep}), machine translation (\citealt{sutskever2014sequence}), and computational biology (\citealt{angermueller2016deep}).  Methods based on Convolutional Neural Networks (CNNs) (\citealt{lecun1998gradient}) and Recurrent Neural Networks (RNNs) (\citealt{bullinaria2013recurrent}) have been proposed for the task of identifying protein binding sites in DNA and RNA sequences, and have achieved state-of-the-art performance (\citealt{alipanahi2015predicting,quang2016danq,hassanzadeh2016deeperbind,shen2018recurrent}).

DeepBind (\citealt{alipanahi2015predicting}) was the first deep learning approach for this task, and used a single layer of convolution and demonstrated the accuracy of these models, as well as their ability to learn signal detectors that recapitulate known motifs.
The work of \citealt{zeng2016convolutional} further showed the value of CNNs and explored in more detail the effect of various architecture parameters such as the number of layers and operations such as pooling.
Other studies opted for more complex architectures and introduced hybrid models that integrate both CNNs and RNNs.
DeeperBind (\citealt{hassanzadeh2016deeperbind}) and DanQ (\citealt{quang2016danq}) for example, add Long Short-Term Memory (LSTM) layer(s) to the DeepBind architecture. The additional RNN layers are designed to improve binding accuracy prediction by learning long-range dependencies between the sequence features learned by the CNN layers.
Purely RNN-based methods were also examined:  the KEGRU method (\citealt{shen2018recurrent}) used a layer of bidirectional Gated Recurrent Units (bi-GRUs), combined with a k-mer embedding representation of the input sequence to create an internal state of the network that allows it to capture long range dependencies and thus obtain good performance.
Methods that are specific to RBP binding were also developed.  For example, 
iDeepS which uses both CNN and RNN layers, identifies sequence and structural motifs simultaneously (\citealt{pan2018prediction}).

Despite all these studies, it is still not clear which deep learning architecture performs best for detecting binding in DNA and RNA sequences.
A fair and unbiased comparison can be very challenging due to many factors including the sensitivity of deep learning methods to the step of model selection  (\citealt{lipton2018troubling}):  deep neural networks have many hyper-parameters that require careful tuning, and differences in performance can be the result of the use of different model selection strategies.
Therefore, a meaningful comparison requires the use of a coherent model selection strategy applied uniformly across all architectures. 
In this study, we conduct a systematic exploration of the performance of different architectures using CNNs and/or RNNs for the study of DNA and RNA sequence binding specificity prediction. 
For this purpose, we have designed a collection of different architecture variants, some of which correspond to published methods, by varying the network components, depth, and input layer representation. 
To ensure the objectivity of our evaluation, we used the same model selection strategy and made the pipeline fully automatic to avoid the need for any hand-tuning and thus remove any bias.

Our experiments use datasets collected from the Encyclopedia of DNA Elements (ENCODE) project (\citealt{encode2004encode}) and verified binding site of RNA binding proteins (RBPs)  derived from large-scale CLIP-seq experiments (\citealt{stravzar2016orthogonal}). 
We find that more complex architectures that combine RNNs and CNNs indeed provide improved performance over the vanilla CNN model, and that this advantage increases with increasing number of training examples that are available. 
However, the improvement in accuracy comes at the expense of the interpretability of the learned models and increased training times.
Our results also demonstrate the advantage of using a k-mer embedding to represent the input sequence instead of the standard one-hot encoding, especially for RBP binding site prediction. 
Finally, We present an end-to-end deep learning toolkit called \deepRAM{} that provides a framework for training and evaluating deep learning architectures for DNA/RNA sequence analysis.


\begin{methods}
\section{Methods}
In this study, we present a comprehensive evaluation of different deep learning architectures for the task of predicting DNA and RNA protein binding sites. First, we present the benchmark datasets used in our study. Then, we present the architectures used in our experiments. Third, we provide the technical details of the model selection process that we followed to ensure unbiased model comparison. These methods are implemented as an open-source deep learning package called \deepRAM{}, that allows users to evaluate different architectures for predicting DNA and RNA protein binding sites. Finally, We describe our method for extracting motifs from the learned models.

\subsection{Datasets}

The deep learning models are evaluated on data from ChIP-seq and CLIP-seq experiments.
For ChIP-seq data we used data from
83 ChIP-seq experiments from the ENCODE project that assayed binding of diverse transcription factors. 
These datasets were used to evaluate deep learning architectures in (\citealt{alipanahi2015predicting}) and (\citealt{zhou2015predicting}), and we use the same sequences as training/testing examples.
The authors of (\citealt{alipanahi2015predicting}) split the ChIP peak data into three categories A, B and C. A is the set of the top 500 even-numbered peaks. B is the set of the top 500 odd-numbered peaks and C is the set of remaining peaks.  For model training, we use the peaks from the A and C and the peaks from B were used for testing.
Positive examples in this binary classification task consist of 101 bp regions centered around each ChIP-seq peak.
The negative examples were generated by shuffling the positive sequences while matching dinucleotide composition.

We also evaluate the ability of different architectures to identify RNA binding sites. 
We use the same benchmark human dataset used by the developers of iONMF (\citealt{stravzar2016orthogonal}) which consists of 31 CLIP-seq experiments over 19 proteins. The data was obtained from (https://github.com/mstrazar/ionmf);  original data was retrieved from  DoRiNA (\citealt{blin2014dorina})  and iCount  (http://icount.biolab.si/). Positive sites represented nucleotides that were identified as being within clusters of interaction sites derived from CLIP-seq. Negative sites were extracted from  genes not participating in the protein-RNA interaction process in any of the 31 experiments. Each experiment consists of 40,000 examples divided into 30,000 examples for training and 10,000 for model testing and evaluation.

\subsection{Model architectures}

\begin{figure*}[ht]
\begin{center}

\includegraphics[width=\linewidth,height=20cm,keepaspectratio]{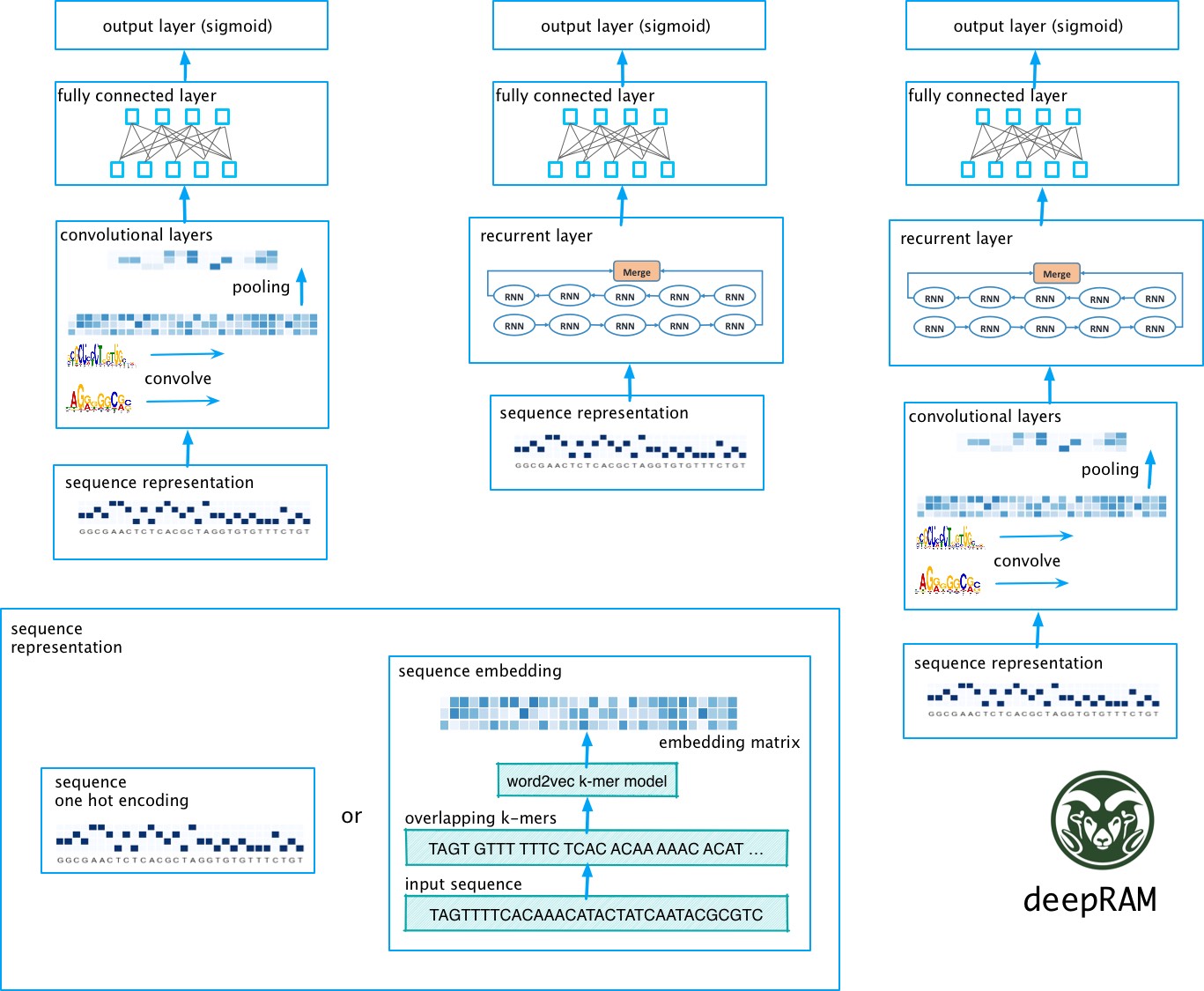}
\caption{Overview of the deep learning architectures evaluated in this work.
These include CNN-only models known for their ability to detect motifs (left), RNN-only models (center) which excel at capturing long-term sequence dependencies, and hybrid CNN-RNN models.
The input for all variants is either a one-hot encoding or a k-mer embedding of the DNA/RNA sequence obtained using word2vec.}
\label{figarch}
\end{center}
\end{figure*}

In this section we describe the variety of deep learning architectures that can be applied to biological sequences (see Figure~\ref{figarch}).
In addition to comparing architectures, we compare two ways of representing the input sequence: either using a one-hot encoding or a k-mer embedding computed using word2vec (\citealt{mikolov2013distributed,asgari2015continuous}). When using the one-hot encoding, the input sequence is represented by a $4 \times L$ matrix where $L$ is the length of the sequence and each position in the sequence has a four element vector with a single nonzero element corresponding to the nucleotide in that position. 
For the The k-mer embedding representation (see Figure~\ref{figarch}), we first split the sequence into overlapping k-mers of length $k$ using a sliding window with stride $s$ and then we map each k-mer in the obtained sequence into d-dimensional vector space using the word2vec algorithm (\citealt{mikolov2013distributed}). 
word2vec is an unsupervised learning algorithm which maps k-mers from the vocabulary to vectors of real numbers in a low-dimensional space. The embedding representation of k-mers is computed in such a way that their context is preserved, i.e. word2vec produces similar embedding vectors for k-mers that tend to co-occur across sequences.


\paragraph{Convolutional Networks.}

CNNs for biological sequence data perform one dimensional convolution:  they slide local signal detectors (filters) along the sequence and integrate their results at increasing spatial scales, generating a representation that is able to abstract away some of the variability observed in binding sites.
Each convolutional module is composed of a convolutional layer and a pooling layer (see Figure~\ref{figarch}). A convolutional layer consists of one-dimensional convolution operation with a specified number of kernels or filters.
The results of applying the filter at each position of the sequence is transformed using a non-linear activation function.  We use the commonly used  rectified linear units (ReLU), which keeps only positive matches and sets the remaining to 0 which helps avoid the vanishing gradient problem. 
More specifically, a convolution layer computes
 
\begin{equation}\label{eq:conv}
    \mathrm{convolution}(X)_{i,k}=\mathrm{ReLU} \left( \sum_{m=0}^{M-1} \sum_{n=0}^{N-1}W_{mn}^k X_{i+m,n} \right),
\end{equation}
where $X$ is the input matrix representing the sequence, $i$ is the index of the output position and $k$ is the index of the filter. Each convolutional filter $W^k$ is an $M \times N$ weight matrix with $M$ being the window size and $N$ being the number of input channels (for the first convolution layer $N$ equals the input representation dimension (4 for one-hot encoding or $d$ for the word2vec representation); for higher-level convolutional layers $N$ is the number of filters in the previous convolutional layer).
Next, the output of convolution undergoes pooling, which aggregates the outputs from neighboring positions
for each filter in order to achieve consistency and invariance to small shifts in the input sequence. 
In this work we use max-pooling which computes the maximum value of a fixed number of spatially adjacent overlapping windows over the convolutional layer's output:

\begin{equation}
    \mathrm{pooling}(Y)_{ik}=\max(Y_{iP,k},Y_{iP+1,k},\dots, Y_{iP+P-1,k}),
\end{equation}
where $Y$ is the output of the convolutional layer, $P$ is the pooling window size, $i$ is the index for output position and $k$ is the index of the filter being pooled.

\begin{table*}[b]
\begin{center}
\processtable{Overview of the models compared in this work. '+' and '-' denote the presence and absence of the layer type respectively. '(.)' denotes the number of convolution layers if present. In the recurrent layers, if present, the type of RNN is specified.\label{Tab:models}}
{\begin{tabular}{@{}cccccccccc@{}}\toprule Layers & DeepBind & DeepBind* & Dilated & DanQ & DanQ* & DeepBind-E* & KEGRU & ECLSTM & ECBLSTM\\\midrule
Embedding &  \textbf{-} & \textbf{-} & \textbf{-} & \textbf{-} & \textbf{-} & \textbf{+} & \textbf{+} & \textbf{+} & \textbf{+}\\
\\
Convolution &  \textbf{+} & \textbf{+}$^{(3)}$ & \textbf{+}$^{(3)*}$ & \textbf{+} & \textbf{+}$^{(3)}$ & \textbf{+}$^{(3)}$ & \textbf{-} & \textbf{+} & \textbf{+} \\
\\
Recurrent&  \textbf{-} & \textbf{-} & \textbf{-} & bi-LSTM & bi-LSTM & \textbf{-} & bi-GRU & LSTM & bi-LSTM\\
\botrule
\end{tabular}}{* The Dilated architecture consists of three convolution layers, one non dilated followed by two dilated (dilation=2) convolution layers}
\end{center}
\end{table*}

The first convolutional layer can be thought of as a motif scanner where each filter is considered as a Position Weight Matrix (PWM) and the convolution operation is equivalent to scanning the PWM with a sliding window across the sequence.
However, the weight matrices associated with convolutional filters are not required to be log-odd ratios. Additional layers of convolution and pooling enable the network to extract features from larger spatial ranges such as motif interactions, which allows it to represent more complex patterns than shallower networks.
Deeper networks have more parameters and require more data for obtaining high levels of performance.

\paragraph{RNN-based models.}
The second class of architectures we explored are RNN-only models.
RNNs have an internal state that is updated as the network progresses along the input sequence.
This internal memory allows RNNs to capture interactions between distant elements along the sequence, and are therefore commonly used in natural language processing (\citealt{hirschberg2015advances}).
Two types of RNN units were tested using \deepRAM{}: LSTM units (\citealt{hochreiter1997long}) and GRU units (\citealt{cho2014properties,chung2014empirical}).
A GRU unit given an input $x_t$ at position $t$ in the sequence performs the following operations:
\begin{equation}\label{eqlstm}
\begin{split} 
z_t &=\sigma (W_z \times [h_{t-1}, x_t] + b_z),\\
r_t &=\sigma (W_r \times [h_{t-1}, x_t] + b_r),\\
\tilde{h}_t &=\tanh(W_h \times[r_t \odot h_{t-1},x_t] + b_h),\\
h_t &= (1- z_t)\odot h_{t-1} + z_t \odot \tilde{h}_t,\\
\end{split}
\end{equation}
where $\odot$ is element-wise multiplication, $z_t$ and $r_t$ are the two GRU gates called the update gate and reset gate, respectively, $W_z$, $W_r$ and $W_h$ are weight matrices, and $b_z$, $b_r$ and $b_h$ are the biases.
$h_t$ is the hidden state which is used as memory to hold information on previous data the network has seen before, $\tilde{h}_t$ is the candidate memory state which is considered to potentially overwrite $h_t$. The reset gate controls how much past information to forget and the update gate controls how much information to throw away and what new information to add. The gates and hidden states are vectors of real numbers of the same dimension, where the dimension is a tunable hyper-parameter.

LSTM units are more complex than GRU units, and we refer the readers to the original publications for details (\citealt{hochreiter1997long}). 
The basic idea of using a gating mechanism in both LSTM and GRU architectures is to capture short term and long term dependencies in sequences. After the LSTM/GRU cell has iterated over the sequence, we output its hidden state at the last position which contains information about the entire sequence. 

bi-RNN (bi-GRU/bi-LSTM) is an extension of the regular RNN which consists of a forward layer and a backward layer representing the positive and negative directions respectively. The forward layer is similar to a regular RNN layer run on the input sequence and the backward layer is another separate RNN layer run on the reverse of the input sequence. The output of the bi-RNN is then computed by concatenating the output vectors of the two layers together. 

\paragraph{Hybrid models}
The third variant in Figure~\ref{figarch}.A are hybrid convolutional and recurrent deep neural networks. The convolution stage which is composed of one or more convolutional modules scans the sequence representation using a set of one-dimensional convolutional filters in order to capture sequence patterns or motifs. The convolutional stage is followed by an RNN stage which is capable of learning complex high-level grammar-like relationships by considering the orientations and spatial distances between the motifs.

The final module in all three types of models is composed of one or two fully connected layers to integrate information from the entire sequence followed by a sigmoid layer to compute the probability that the input sequence contains a DNA- or RNA-binding binding site.

\paragraph{Evaluated architectures}
The \deepRAM{} tool provides implementations of several existing architectures:
DeepBind (\citealt{alipanahi2015predicting}) which uses a single-layer CNN layer, DanQ which uses a single-layer CNN and bidirectional LSTM (\citealt{quang2016danq}), KEGRU which uses k-mer embedding and GRU units (\citealt{shen2018recurrent}) and dilated multi-layer CNN (\citealt{gupta2017dilated}). 
To fully evaluate the range of deep learning architectures we considered additional variants
denoted as DeepBind* (multi-layer CNN), DanQ* (DanQ with multiple layers of convolution), DeepBind-E* (multi-layer CNN with k-mer embedding), ECLSTM (k-mer embedding with single layer CNN and LSTM) and ECBLSTM (k-mer embedding with single layer CNN and bi-directional LSTM). 
These architectures are summarized in Table~\ref{Tab:models}. 

\subsection{Model training, selection, and evaluation}

Model Selection is perhaps the most challenging step in deep learning as the performance of deep learning algorithms is very sensitive to the calibration parameters (\citealt{lipton2018troubling}).  A careful configuration and selection of the hyper-parameters is thus essential.  For each dataset, we use automatic calibration that is based on randomly sampling 40 hyper-parameter settings from all possible combinations; for each setting, a model is trained using 3-fold cross-validation.
We use the area under the ROC curve (AUC) to evaluate the performance of the model and each calibration set is scored by its average AUC in 3-fold cross-validation. 
Next, we use the selected best hyper-parameter set to train five new models using the full training data to avoid random initialization effects and then choose the model with the best training performance as the final selected model that is then used for prediction of sequences in the test set.  This model selection strategy is based on the one used by the authors of DeepBind (\citealt{alipanahi2015predicting}).

In the training phase, we consider the number of learning steps as a hyper-parameter. For each of the 40 calibration sets, we train a model for a maximum of 40,000 learning steps and test it on the held out validation set every 5,000 learning steps. The iteration with the best validation accuracy is picked as the number of learning steps in which the model performed best on validation. The selected number of learning steps is added to the calibration set as an additional hyper-parameter. We select the iteration with the best validation score because we assume that the model starts to over-fit after the selected iteration.
The number of filters in the first convolutional layer is chosen as part of model selection; the number of filters in each subsequent layer is increased by 50\% compared to the layer before it.

\paragraph{Model training.}

To train a given model, we minimize the cross-entropy objective function.
Derivatives of the objective function with respect to the model parameters were computed by back-propagation.  Minimizing the objective function is performed by Stochastic Gradient Descent (SGD) or Adagrad, and the choice is made as part of the model selection process.
Examples were processed using a batch size of 128 in all experiments.
We used multiple regularization schemes including dropout (applied to max pooling layers/RNN layers/hidden layers), weight decay, and early stopping.
Details of the hyper-parameter space are summarized in Table~\ref{Tab:hyper}.

We ran our experiments on an Ubuntu server with a TITAN X GPU
with 12 GB of memory.
Typical running times of each experiment for model selection is between one hour for a single layer CNN to almost four hours for a network that includes convolutional and bi-LSTM modules (see details in Table S3 in the supplementary file).

\begin{table}[!ht]
\processtable{\deepRAM{} Hyper-parameters, search space and sampling method.
\label{Tab:hyper}} 
{\begin{tabular}{@{}llll@{}}\toprule 
Calibration Parameters & Search Space & Sampling \\\midrule
Embedding size & 50 & Fixed\\
Embedding k-mer length & 3 & Fixed \\
Embedding stride & 1 & Fixed \\
motif length & \{10 , 24 \}* & Fixed\\
Number of filters & \{16, 32\} & uniform \\
Pooling window size & 3 & Fixed \\
Pooling stride & 1 & Fixed \\
RNN hidden size & \{20,50,80,100\} & uniform \\
Neural Net hidden layer & \{Nan, 32units, 64units\} & uniform \\
optimizer & \{SGD, Adagrad\} & uniform \\
learning rate & [1e-3,1e-1] & log uniform \\
learning momentum(SGD) & [0.95,0.99] & sqrt uniform \\
number of learning steps & [5,000:40,000]** & evaluate all \\
weight initialization & \{xavier, normal\} & uniform \\
initial weight scale(motifs) & [1e-6,1e-1] & log uniform \\
initial weight scale (RNN) & [1e-6,1e-1] & log uniform \\
initial weight scale (NN) & [1e-5,1e-1] & log uniform \\
weight decay & [1e-10,1e-1] & log uniform \\
dropout expectation & \{0.4, 0.55, 0.7, 0.85, 1\} & uniform \\\botrule
\end{tabular}}{ * 10 with k-mer embedding. 24 with one-hot encoding \\
** step= 5000}
\end{table}

\subsection{Motif extraction}

In order to make models implemented using \deepRAM{} easily interpretable, we extract motifs from the first convolutional layer following a similar methodology as in DeepBind (\citealt{alipanahi2015predicting}).
To do so, we feed all test sequences through the convolution stage.  
For each filter, we extract all sequence fragments that activate the filter and use only activations that are greater than half of the filter's maximum value over all sequences. 
Once all the sequence fragments are extracted, they are stacked and the nucleotide frequencies are counted to form a  position frequency matrix (PFM).  Sequence logos are then constructed using WebLogo (\citealt{crooks2004weblogo}). 
Finally, these discovered motifs are aligned using TOMTOM (\citealt{gupta2007quantifying}) against known motifs from CISBP-RNA (\citealt{ray2013compendium}) for RBPs and JASPAR (\citealt{mathelier2013jaspar}) for transcription factors. 

\subsection{\deepRAM{}}

\deepRAM{} is an end-to-end deep learning toolkit for predicting protein binding sites and motifs. It helps users run experiments using many state-of-the-art methods
and addresses the challenge of selecting model parameters in deep learning models
using a fully automatic model selection strategy.
This helps avoid hand-tuning and thus removes any bias in running experiments, making it user friendly without losing its flexibility. 
While it was designed with ChIP-seq and CLIP-seq data in mind, it can be used for any DNA/RNA sequence binary classification problem.

\deepRAM{} allows users the flexibility to choose a deep learning model by selecting its different components:  input sequence representation (one-hot or k-mer embedding), whether to use a CNN and how many layers, and whether to use an RNN, and the number of layers and their type.
For CNNs the user can choose to use dilated convolution as well.
Once the model is trained, the learned motifs of the first convolutional module are automatically extracted and visualized using Weblogo, and then matched with known motifs using TOMTOM.

We implemented \deepRAM{} using PyTorch 1.0 (http://pytorch.org/), which supports GPU acceleration. Our implementation has been packaged to make it
runnable on any Unix-based system, and is available at:
\href{https://github.com/MedChaabane/deepRAM}{https://github.com/MedChaabane/deepRAM}.

\end{methods}

\section{Results}

\subsection{Deeper is better}
\begin{figure*}[t]
\centering
\includegraphics[width=\textwidth]{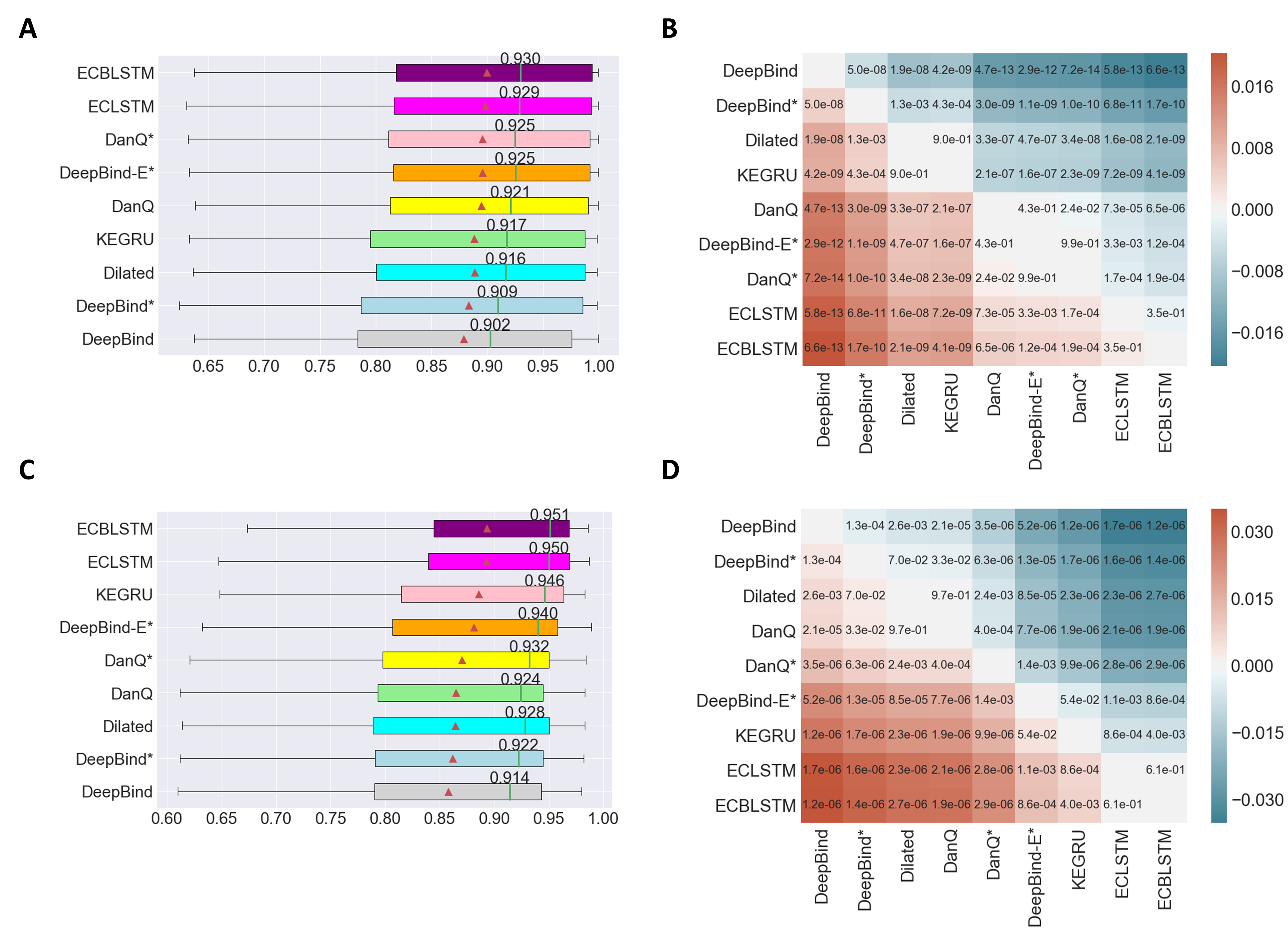}
\caption{(A) The distribution of AUCs across 83 ChIP-seq datasets. 
(B) Heatmap annotated with p-values of pairwise model comparison using the Wilcoxon signed-rank test for ChIP-seq datasets. 
(C) The distribution of AUCs across 31 datasets for predicting RBP binding sites. 
(D) Heatmap annotated with p-values of pairwise model comparison using the Wilcoxon signed-rank test for predicting RBP binding sites. 
In subfigures (A) and (C), the triangle represents the average AUC for the respective model, the annotated vertical line represents the median AUC whose value is indicated.
The models are sorted by their average AUC values. 
In subfigures (B) and (D), the color red or blue at position $(i,j)$ in the heatmap indicates which model has a high average AUC, and its intensity indicates the magnitude of the difference.}
\label{figrslt}

\end{figure*}

We evaluate and compare the performance of the different models discussed in section 2.2 on the two tasks of predicting DNA- and RNA- protein binding sites (see Figure~\ref{figrslt}). Overall, all models perform well with all median AUCs greater than 0.90 on ChIP-seq data and greater than 0.91 on CLIP-seq data. The proposed ECBLSTM model (Embedding, Convolution, bi-LSTM) provides the most significant improvement over DeepBind with a median AUC of 0.930 compared with 0.902 for DeepBind on ChIP-seq data, and with a more pronounced gap for CLIP-seq data:  0.951 for ECBLSTM vs 0.914 for DeepBind.
All the performance differences described here are statistically significant except when noted explicitly (see Figure~\ref{figrslt}).
Detailed accuracy values for individual datasets are provided in Tables S1 and S2 in the supplementary file.

DeepBind is the simplest model considered here:  it uses one-hot sequence encoding, and a single convolutional layer.
The results shown in Figure~\ref{figrslt} demonstrate that adding multiple convolutional layers, dilated convolution, and sequence embedding 
all provide improved performance over the original DeepBind.
The addition of a recurrent module provides further improvement as seen by 
comparing the performance of ECBLSTM to a model called DeepBind-E* which has multiple convolutional layers and an embedding stage.
This shows that adding recurrent connections to capture long-term dependencies between motifs detected by the convolutional layer leads to improved performance.
The performance advantage of RNNs is further highlighted by comparing the performance of DanQ where the additional bi-directional LSTM layer has helped improve its performance over DeepBind.

We note that iDeepS which is specifically designed for RNA binding and uses a CNN over sequence and local secondary structure in combination with an LSTM module, achieved a median AUC of 0.917 for the CLIP-seq data, which is less than all the evaluated methods except DeepBind (see Table S4 in the supplementary file).
All the deep learning methods performed better than 
iONMF which uses multiple sources of data, including k-mer frequency, secondary structure, GO annotations (see Table S4 in the supplementary file).

We note that in both tasks, our implementation of DeepBind achieved nearly identical performance to the original DeepBind implementation (see Figure~S3 in the supplementary file). 

\subsection{k-mer embedding boosts model performance}

We observe that using k-mer embedding to represent input sequences rather than one-hot encoding improves model performance, and more so for the RBP binding datasets. 
For example, among models with the same architecture, we see that ECBLSTM outperforms DanQ in both tasks (see Figure~\ref{figrslt} and Supplementary Figures S2 and S3). 
We also observe that in the task of RNA-protein binding site prediction, all models that use embedding representation have median AUC higher than 0.94 while all models that use one-hot encoding have median AUC lower than 0.935 (Figure~\ref{figrslt}.C). 
These results suggest that one-hot encoding is perhaps not the optimal strategy for representation of DNA and RNA sequences. In contrast, k-mer embedding integrates the contextual information of k-mers by learning the statistical information of k-mer co-occurrence relationships in the input sequences.

In this work, we train the k-mer embedding algorithm for each dataset with $k=3$ and stride $s=1$.
Other studies (\citealt{shen2018recurrent,min2017chromatin}) have encouraged the use of larger values of stride and k-mer length and suggested that the use of small stride values ($s=1$) may affect negatively the performance of the embedding algorithm. 
In preliminary experiments, we found that using small values of stride and k-mer length ($k=3$ and $s=1$) has the best performance.

\subsection{Deeper is better with sufficient training data}

\begin{figure*}[ht]
\centering
\includegraphics[width=\linewidth]{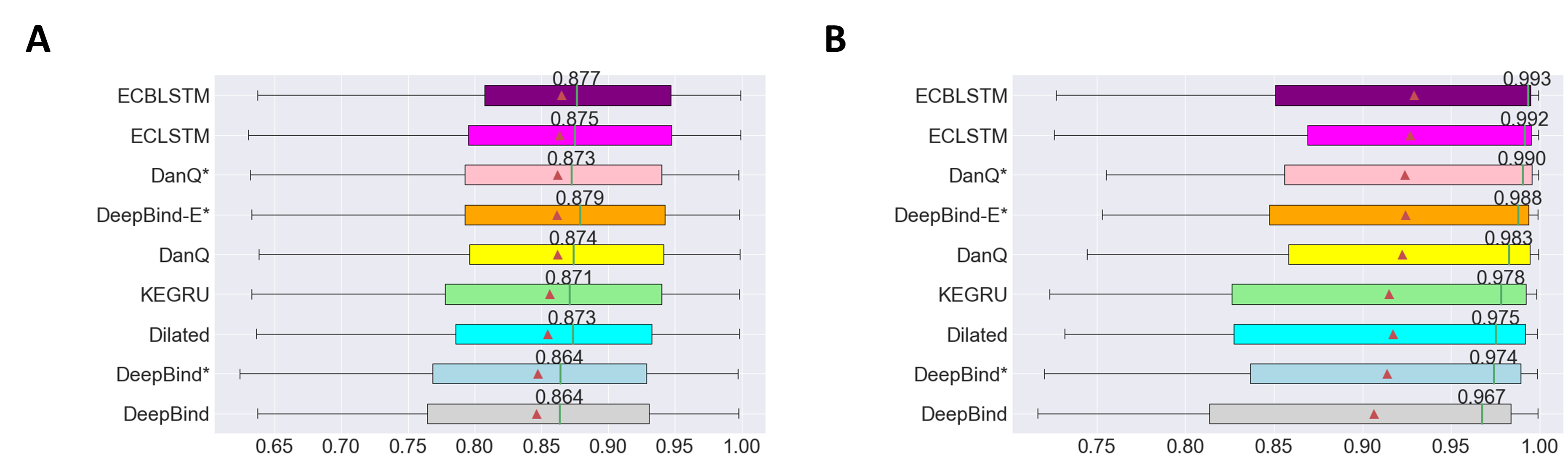}
\caption{(A) The distribution of AUCs in predicting DNA protein binding sites across 38 ChIP-seq experiments with less than 10,000 peaks . (B)  The distribution of AUCs in predicting DNA protein binding sites across 45 ChIP-seq experiments with more than 10,000 peaks.  Figure notation follows the description in Figure~\ref{figrslt}.}
\label{figboxes}
\end{figure*}

Based on the results shown in Figure~\ref{figrslt}, one may conclude that relatively complex models tend to perform better than simpler models. 
However, this statement is based on the evaluation of the overall performance across all experiments and do not take into consideration the effect of the number of training examples.
To study this aspect, we divided the ENCODE ChIP-seq datasets into two groups according to the number of training examples.  The first group consists of 38 datasets with less than 10,000 positive training samples, and the second group consists of 45 datasets with more than 10,000 positive training samples. 
We compare the performance of different models in these two groups and report the results in Figure~\ref{figboxes}. 
We observe considerably higher AUCs for the large datasets with median AUCs between 0.967 (DeepBind) and 0.993 (ECBLSTM) compared to median AUCs between 0.864 (DeepBind) and 0.879 (DeepBind-E*) for the small datasets.
It is also worth noting that the effect of the number of training examples is more pronounced with hybrid models (see Figure~\ref{figboxes} and Supplementary Figure~S4). Indeed, ECBLSTM, ECLSTM and DanQ* tend to perform very strongly for large datasets (median AUCs above 0.983) while interestingly, they fell behind DeepBind-E* when used on smaller datasets.
this suggests the need  for sufficient training data for hybrid models.
Complex models such as ECBLSTM still perform well even for the smaller datasets, demonstrating that our regularization procedure was effective in preventing over-fitting.


\subsection{Dilated convolution}

Dilated convolution uses filters with gaps to allow each filter to capture information across larger and larger stretches of the input sequence (\citealt{yu2015multi}). Hence, dilated convolution finds usage in applications that benefit from modeling of a wider context without incurring the increased cost of using RNNs (\citealt{gupta2017dilated,strubell2017fast,kelley2018sequential}). 

In this work, we evaluate a dilated model which consists of three convolutional modules with dilations equal to 1, 2 and 2 in the first, second and third layers, respectively. We find that dilated convolutional model outperforms DeepBind* with significant p-values in both tasks (Figure~\ref{figrslt}).
In addition, the dilated convolutional model had slightly higher median AUC than DanQ in the RBP binding sites datasets, which suggests that dilated convolution can capture long range relationships similarly to LSTMs.
These findings suggest that dilated convolution are a valuable architecture parameter
to consider.
This is likely to be even more pronounced for longer sequences such as those modeled using the Basenji method (\citealt{kelley2018sequential}).

\subsection{Model interpretation and visualization}

\begin{figure*}[t]
\begin{center}

\includegraphics[width=\linewidth,height=20cm,keepaspectratio]{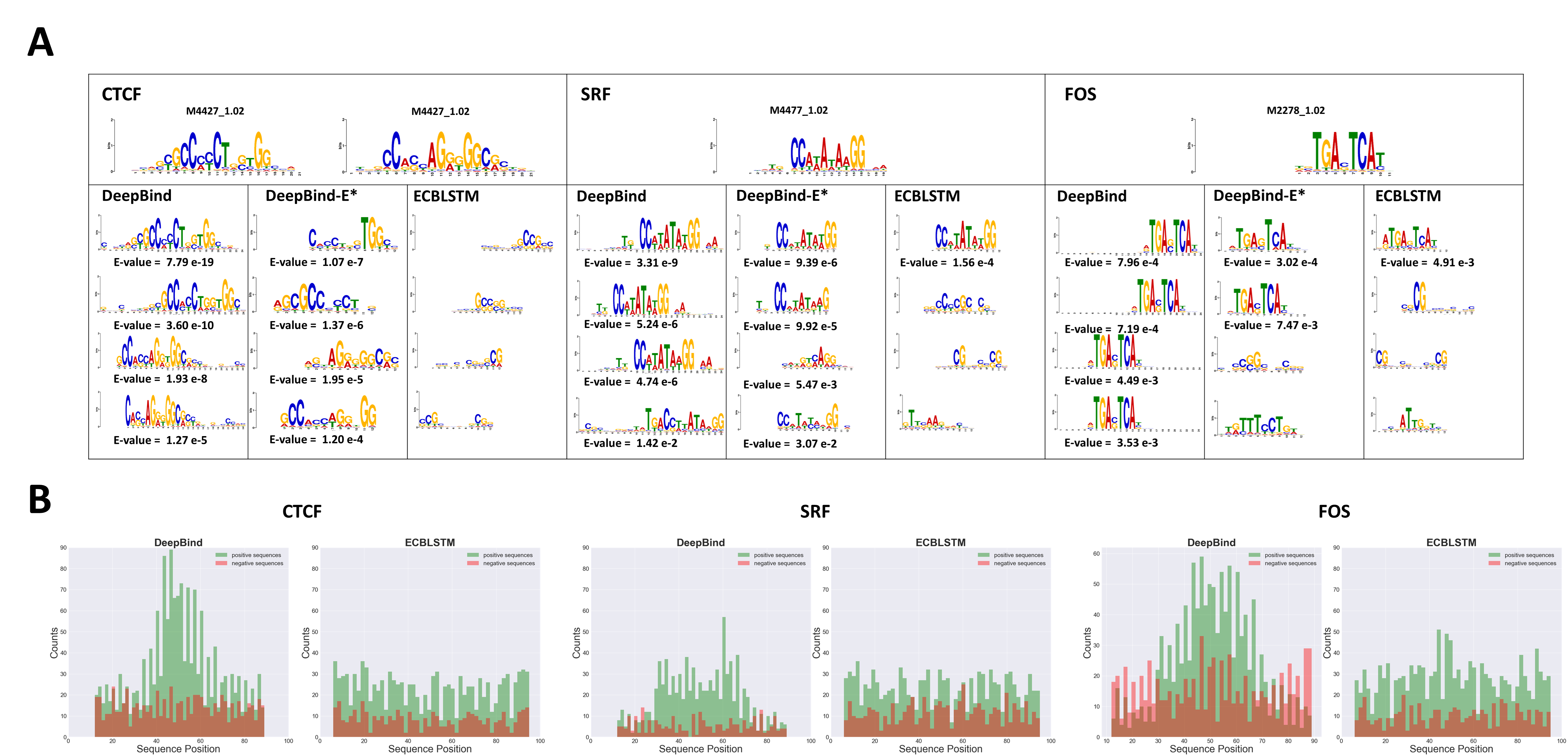}
\caption{(A) Examples of motifs detected by first layer convolutional modules learned by DeepBind, DeepBind-E* and ECBLSTM for predicting DNA binding sites of CTCF, SRF, and FOS. E-values are displayed below each motif only if it matches with the known motifs for the same Transcription Factor. Known motifs from the JASPAR database are displayed at the top. 
(B) Histograms of the counts of convolutional filter activations that are considered for extracting motifs along the sequences for models of CTCF, SRF and FOS binding using DeepBind and ECBLSTM.}
\label{fig5}
\end{center}
\end{figure*}

To explore the ability of selected architectures to capture informative motifs, we converted filters of the first convolutional layer to sequence motifs as described in section 2.4. As shown in Figure~\ref{fig5}.A, DeepBind and DeepBind-E* are able to detect informative motifs that match well with known motifs from the JASPAR database. However, ECBLSTM turns out to perform poorly in detecting motifs compared to the two other models and most of its detected motifs are not informative despite the fact that it is the best performing model among all the models we compared.
We hypothesize that when combined with RNNs, the CNN filters learn information that is geared towards providing the subsequent recurrent layer with the information it needs, which is of a different nature than the localized information learned by CNN-only models.

To further investigate the difference between the behaviour of hybrid models and CNN-only models, We explored the distribution of sequence fragments with positive activation values for a given filter with DeepBind and ECBLSTM in the positive and negative examples (Figure~\ref{fig5}.B ). As expected, the number of activated sequence fragments in positive sequences is much higher than in negative sequences in both methods. 
In addition, We observe that the activated sequence fragments in positive sequences using DeepBind are concentrated in the middle of the sequence, and are uniformly distributed for negative examples.
However, using ECBLSTM the activated sequence fragments are distributed uniformly across the sequence for both positive and negative sequences. Noting that the centers of positive sequences correspond to the reported ChIP-seq peaks, we conclude that DeepBind is detecting sequence motifs that represent the binding event while ECBLSTM's convolution stage is extracting features that span the whole sequence.
This is in agreement with our finding that RNNs lead to a representation which has reduced interpretability compared to that of CNNs.

%
%

\section{Conclusion}

In this work we performed a thorough analysis and evaluation of the performance of commonly used deep learning architectures for DNA and RNA binding site prediction.
This study aims at helping researchers to get a better understanding of the performance characteristics and advantages
of different architectures to help them choose the right architecture for their work.
Our experiments demonstrated the accuracy of hybrid CNN/RNN models;
however, that requires the availability of sufficient training data, and these networks are harder to interpret and hence their usefulness in motif discovery might be limited.  
We have made the software used in our experiments available as an easy to use
tool to evaluate and analyze various deep learning architectures for DNA/RNA binding prediction in a user-friendly package called \deepRAM{}. 
We hope this work will stimulate further studies on visualizing and understanding deep models and enhance their usefulness for analyzing biological sequence data.




\bibliographystyle{natbib}
\bibliographystyle{achemnat}
\bibliographystyle{plainnat}
\bibliographystyle{abbrv}
\bibliographystyle{bioinformatics}

\bibliographystyle{plain}

\bibliography{document.bib}

\begin{thebibliography}{}

\bibitem[Alipanahi {\em et~al.}(2015)Alipanahi, Delong, Weirauch, and
  Frey]{alipanahi2015predicting}
Alipanahi, B., Delong, A., Weirauch, M.~T., and Frey, B.~J. (2015).
\newblock Predicting the sequence specificities of {DNA}-and {RNA}-binding
  proteins by deep learning.
\newblock {\em Nature biotechnology\/}, {\bf 33}(8), 831.

\bibitem[Angermueller {\em et~al.}(2016)Angermueller, P{\"a}rnamaa, Parts, and
  Stegle]{angermueller2016deep}
Angermueller, C., P{\"a}rnamaa, T., Parts, L., and Stegle, O. (2016).
\newblock Deep learning for computational biology.
\newblock {\em Molecular systems biology\/}, {\bf 12}(7), 878.

\bibitem[Asgari and Mofrad(2015)Asgari and Mofrad]{asgari2015continuous}
Asgari, E. and Mofrad, M.~R. (2015).
\newblock Continuous distributed representation of biological sequences for
  deep proteomics and genomics.
\newblock {\em PloS one\/}, {\bf 10}(11), e0141287.

\bibitem[Blin {\em et~al.}(2014)Blin, Dieterich, Wurmus, Rajewsky, Landthaler,
  and Akalin]{blin2014dorina}
Blin, K., Dieterich, C., Wurmus, R., Rajewsky, N., Landthaler, M., and Akalin,
  A. (2014).
\newblock Dorina 2.0--upgrading the {doRiNA} database of {RNA} interactions in
  post-transcriptional regulation.
\newblock {\em Nucleic acids research\/}, {\bf 43}(D1), D160--D167.

\bibitem[Bullinaria(2013)Bullinaria]{bullinaria2013recurrent}
Bullinaria, J.~A. (2013).
\newblock Recurrent neural networks.
\newblock {\em Neural Computation: Lecture\/}, {\bf 12}.

\bibitem[Cho {\em et~al.}(2014)Cho, Van~Merri{\"e}nboer, Bahdanau, and
  Bengio]{cho2014properties}
Cho, K., Van~Merri{\"e}nboer, B., Bahdanau, D., and Bengio, Y. (2014).
\newblock On the properties of neural machine translation: {Encoder-decoder}
  approaches.
\newblock {\em arXiv preprint arXiv:1409.1259\/}.

\bibitem[Chung {\em et~al.}(2014)Chung, Gulcehre, Cho, and
  Bengio]{chung2014empirical}
Chung, J., Gulcehre, C., Cho, K., and Bengio, Y. (2014).
\newblock Empirical evaluation of gated recurrent neural networks on sequence
  modeling.
\newblock {\em arXiv preprint arXiv:1412.3555\/}.

\bibitem[Consortium(2004)Consortium]{encode2004encode}
Consortium, E.~P. (2004).
\newblock The {ENCODE (ENCyclopedia of DNA elements)} project.
\newblock {\em Science\/}, {\bf 306}(5696), 636--640.

\bibitem[Crooks {\em et~al.}(2004)Crooks, Hon, Chandonia, and
  Brenner]{crooks2004weblogo}
Crooks, G.~E., Hon, G., Chandonia, J.-M., and Brenner, S.~E. (2004).
\newblock {WebLogo}: a sequence logo generator.
\newblock {\em Genome research\/}, {\bf 14}(6), 1188--1190.

\bibitem[Ferr{\'e} {\em et~al.}(2016)Ferr{\'e}, Colantoni, and
  Helmer-Citterich]{Ferre}
Ferr{\'e}, F., Colantoni, A., and Helmer-Citterich, M. (2016).
\newblock Revealing protein-{lncRNA} interaction.
\newblock {\em Briefings in Bioinformatics\/}, {\bf 17}(1), 106--116.

\bibitem[Gerstberger {\em et~al.}(2014)Gerstberger, Hafner, and
  Tuschl]{gerstberger}
Gerstberger, S., Hafner, M., and Tuschl, T. (2014).
\newblock A census of human {RNA}-binding proteins.
\newblock {\em Nature Reviews Genetics\/}, {\bf 15}(12), 829.

\bibitem[Gupta and Rush(2017)Gupta and Rush]{gupta2017dilated}
Gupta, A. and Rush, A.~M. (2017).
\newblock {Dilated Convolutions for Modeling Long-Distance Genomic
  Dependencies}.
\newblock {\em arXiv preprint arXiv:1710.01278\/}.

\bibitem[Gupta {\em et~al.}(2007)Gupta, Stamatoyannopoulos, Bailey, and
  Noble]{gupta2007quantifying}
Gupta, S., Stamatoyannopoulos, J.~A., Bailey, T.~L., and Noble, W.~S. (2007).
\newblock Quantifying similarity between motifs.
\newblock {\em Genome biology\/}, {\bf 8}(2), R24.

\bibitem[Hassanzadeh and Wang(2016)Hassanzadeh and
  Wang]{hassanzadeh2016deeperbind}
Hassanzadeh, H.~R. and Wang, M.~D. (2016).
\newblock {DeeperBind}: Enhancing prediction of sequence specificities of {DNA}
  binding proteins.
\newblock In {\em Bioinformatics and Biomedicine (BIBM), 2016 IEEE
  International Conference on\/}, pages 178--183. IEEE.

\bibitem[Hinton {\em et~al.}(2012)Hinton, Deng, Yu, Dahl, Mohamed, Jaitly,
  Senior, Vanhoucke, Nguyen, Sainath, {\em et~al.}]{hinton2012deep}
Hinton, G., Deng, L., Yu, D., Dahl, G.~E., Mohamed, A.-r., Jaitly, N., Senior,
  A., Vanhoucke, V., Nguyen, P., Sainath, T.~N., {\em et~al.} (2012).
\newblock Deep neural networks for acoustic modeling in speech recognition: The
  shared views of four research groups.
\newblock {\em IEEE Signal processing magazine\/}, {\bf 29}(6), 82--97.

\bibitem[Hirschberg and Manning(2015)Hirschberg and
  Manning]{hirschberg2015advances}
Hirschberg, J. and Manning, C.~D. (2015).
\newblock Advances in natural language processing.
\newblock {\em Science\/}, {\bf 349}(6245), 261--266.

\bibitem[Hochreiter and Schmidhuber(1997)Hochreiter and
  Schmidhuber]{hochreiter1997long}
Hochreiter, S. and Schmidhuber, J. (1997).
\newblock Long short-term memory.
\newblock {\em Neural computation\/}, {\bf 9}(8), 1735--1780.

\bibitem[Kazan {\em et~al.}(2010)Kazan, Ray, Chan, Hughes, and
  Morris]{kazan2010rnacontext}
Kazan, H., Ray, D., Chan, E.~T., Hughes, T.~R., and Morris, Q. (2010).
\newblock {RNAcontext}: a new method for learning the sequence and structure
  binding preferences of {RNA}-binding proteins.
\newblock {\em {PLoS} computational biology\/}, {\bf 6}(7), e1000832.

\bibitem[Kelley {\em et~al.}(2018)Kelley, Reshef, Bileschi, Belanger, McLean,
  and Snoek]{kelley2018sequential}
Kelley, D.~R., Reshef, Y., Bileschi, M., Belanger, D., McLean, C.~Y., and
  Snoek, J. (2018).
\newblock Sequential regulatory activity prediction across chromosomes with
  convolutional neural networks.
\newblock {\em Genome research\/}, pages gr--227819.

\bibitem[Krizhevsky {\em et~al.}(2012)Krizhevsky, Sutskever, and
  Hinton]{krizhevsky2012imagenet}
Krizhevsky, A., Sutskever, I., and Hinton, G.~E. (2012).
\newblock Imagenet classification with deep convolutional neural networks.
\newblock In {\em Advances in neural information processing systems\/}, pages
  1097--1105.

\bibitem[LeCun {\em et~al.}(1998)LeCun, Bottou, Bengio, and
  Haffner]{lecun1998gradient}
LeCun, Y., Bottou, L., Bengio, Y., and Haffner, P. (1998).
\newblock Gradient-based learning applied to document recognition.
\newblock {\em Proceedings of the IEEE\/}, {\bf 86}(11), 2278--2324.

\bibitem[LeCun {\em et~al.}(2015)LeCun, Bengio, and Hinton]{lecun2015deep}
LeCun, Y., Bengio, Y., and Hinton, G. (2015).
\newblock Deep learning.
\newblock {\em nature\/}, {\bf 521}(7553), 436.

\bibitem[Lipton and Steinhardt(2018)Lipton and Steinhardt]{lipton2018troubling}
Lipton, Z.~C. and Steinhardt, J. (2018).
\newblock Troubling trends in machine learning scholarship.
\newblock {\em arXiv preprint arXiv:1807.03341\/}.

\bibitem[Mathelier {\em et~al.}(2013)Mathelier, Zhao, Zhang, Parcy,
  Worsley-Hunt, Arenillas, Buchman, Chen, Chou, Ienasescu, {\em
  et~al.}]{mathelier2013jaspar}
Mathelier, A., Zhao, X., Zhang, A.~W., Parcy, F., Worsley-Hunt, R., Arenillas,
  D.~J., Buchman, S., Chen, C.-y., Chou, A., Ienasescu, H., {\em et~al.}
  (2013).
\newblock {JASPAR} 2014: an extensively expanded and updated open-access
  database of transcription factor binding profiles.
\newblock {\em Nucleic acids research\/}, {\bf 42}(D1), D142--D147.

\bibitem[Mikolov {\em et~al.}(2013)Mikolov, Sutskever, Chen, Corrado, and
  Dean]{mikolov2013distributed}
Mikolov, T., Sutskever, I., Chen, K., Corrado, G.~S., and Dean, J. (2013).
\newblock Distributed representations of words and phrases and their
  compositionality.
\newblock In {\em Advances in neural information processing systems\/}, pages
  3111--3119.

\bibitem[Min {\em et~al.}(2017)Min, Zeng, Chen, Chen, and
  Jiang]{min2017chromatin}
Min, X., Zeng, W., Chen, N., Chen, T., and Jiang, R. (2017).
\newblock Chromatin accessibility prediction via convolutional long short-term
  memory networks with k-mer embedding.
\newblock {\em Bioinformatics\/}, {\bf 33}(14), i92--i101.

\bibitem[Pan {\em et~al.}(2018)Pan, Rijnbeek, Yan, and Shen]{pan2018prediction}
Pan, X., Rijnbeek, P., Yan, J., and Shen, H.-B. (2018).
\newblock Prediction of {RNA}-protein sequence and structure binding
  preferences using deep convolutional and recurrent neural networks.
\newblock {\em BMC genomics\/}, {\bf 19}(1), 511.

\bibitem[Quang and Xie(2016)Quang and Xie]{quang2016danq}
Quang, D. and Xie, X. (2016).
\newblock {DanQ}: a hybrid convolutional and recurrent deep neural network for
  quantifying the function of {DNA} sequences.
\newblock {\em Nucleic acids research\/}, {\bf 44}(11), e107--e107.

\bibitem[Ray {\em et~al.}(2013)Ray, Kazan, Cook, Weirauch, Najafabadi, Li,
  Gueroussov, Albu, Zheng, Yang, {\em et~al.}]{ray2013compendium}
Ray, D., Kazan, H., Cook, K.~B., Weirauch, M.~T., Najafabadi, H.~S., Li, X.,
  Gueroussov, S., Albu, M., Zheng, H., Yang, A., {\em et~al.} (2013).
\newblock A compendium of {RNA}-binding motifs for decoding gene regulation.
\newblock {\em Nature\/}, {\bf 499}(7457), 172.

\bibitem[Rohs {\em et~al.}(2010)Rohs, Jin, West, Joshi, Honig, and
  Mann]{rohs2010origins}
Rohs, R., Jin, X., West, S.~M., Joshi, R., Honig, B., and Mann, R.~S. (2010).
\newblock Origins of specificity in protein-{DNA} recognition.
\newblock {\em Annual review of biochemistry\/}, {\bf 79}, 233--269.

\bibitem[Shen {\em et~al.}(2018)Shen, Bao, and Huang]{shen2018recurrent}
Shen, Z., Bao, W., and Huang, D.-S. (2018).
\newblock {Recurrent Neural Network for Predicting Transcription Factor Binding
  Sites}.
\newblock {\em Scientific reports\/}, {\bf 8}(1), 15270.

\bibitem[Siggers and Gordan(2013)Siggers and Gordan]{siggers2013protein}
Siggers, T. and Gordan, R. (2013).
\newblock Protein--{DNA} binding: complexities and multi-protein codes.
\newblock {\em Nucleic acids research\/}, {\bf 42}(4), 2099--2111.

\bibitem[Stormo(2000)Stormo]{stormo2000dna}
Stormo, G.~D. (2000).
\newblock {DNA} binding sites: representation and discovery.
\newblock {\em Bioinformatics\/}, {\bf 16}(1), 16--23.

\bibitem[Stra{\v{z}}ar {\em et~al.}(2016)Stra{\v{z}}ar, {\v{Z}}itnik, Zupan,
  Ule, and Curk]{stravzar2016orthogonal}
Stra{\v{z}}ar, M., {\v{Z}}itnik, M., Zupan, B., Ule, J., and Curk, T. (2016).
\newblock Orthogonal matrix factorization enables integrative analysis of
  multiple {RNA} binding proteins.
\newblock {\em Bioinformatics\/}, {\bf 32}(10), 1527--1535.

\bibitem[Strubell {\em et~al.}(2017)Strubell, Verga, Belanger, and
  Mccallum]{strubell2017fast}
Strubell, E., Verga, P., Belanger, D., and Mccallum, A. (2017).
\newblock Fast and accurate sequence labeling with iterated dilated
  convolutions.
\newblock {\em CoRR\/}.

\bibitem[Sutskever {\em et~al.}(2014)Sutskever, Vinyals, and
  Le]{sutskever2014sequence}
Sutskever, I., Vinyals, O., and Le, Q.~V. (2014).
\newblock Sequence to sequence learning with neural networks.
\newblock In {\em Advances in neural information processing systems\/}, pages
  3104--3112.

\bibitem[Yu and Koltun(2015)Yu and Koltun]{yu2015multi}
Yu, F. and Koltun, V. (2015).
\newblock Multi-scale context aggregation by dilated convolutions.
\newblock {\em arXiv preprint arXiv:1511.07122\/}.

\bibitem[Zeng {\em et~al.}(2016)Zeng, Edwards, Liu, and
  Gifford]{zeng2016convolutional}
Zeng, H., Edwards, M.~D., Liu, G., and Gifford, D.~K. (2016).
\newblock Convolutional neural network architectures for predicting
  {DNA}--protein binding.
\newblock {\em Bioinformatics\/}, {\bf 32}(12), i121--i127.

\bibitem[Zhou and Troyanskaya(2015)Zhou and Troyanskaya]{zhou2015predicting}
Zhou, J. and Troyanskaya, O.~G. (2015).
\newblock Predicting effects of noncoding variants with deep learning--based
  sequence model.
\newblock {\em Nature methods\/}, {\bf 12}(10), 931.

\end{thebibliography}

\end{document}